%% file: root.tex
\documentclass[letterpaper, 10 pt, conference]{ieeeconf}

\IEEEoverridecommandlockouts                              %

\usepackage{graphics} %
\usepackage{amsmath} %
\usepackage{amssymb}  %
\usepackage{amsfonts}
\usepackage{bbm}
\usepackage{mathtools}
\usepackage{multirow}
\usepackage{units}
\usepackage{bm}
\usepackage{xcolor}
\usepackage{hyperref}
\let\labelindent\relax
\usepackage{enumitem}
\usepackage[ruled,vlined]{algorithm2e}
\SetCommentSty{mycommfont} %
\usepackage{cite}
\usepackage{float}
\usepackage{microtype}
\usepackage{eso-pic}

\input{math}

\newcommand{\crel}[1]{%
  \global\setbox1=\hbox{$#1$}%
  \global\dimen1=0.5\wd1
  \mathrel{\hbox to\dimen1{$#1$\hss}}&\mathrel{\mspace{-\thickmuskip}\hbox to\dimen1{}}%
}

\newcommand\Tstrut{\rule{0pt}{2.6ex}}         %
\newcommand\Bstrut{\rule[-0.9ex]{0pt}{0pt}}   %

\title{\LARGE \bf
Fast and Robust Bio-inspired Teach and Repeat Navigation
}

\author{Dominic Dall'Osto, Tobias Fischer and Michael Milford%
\thanks{The authors are with the QUT Centre for Robotics, Queensland University of Technology, Brisbane, QLD 4000, Australia (e-mail: {\footnotesize \href{mailto:dominic.dallosto@qut.edu.au}{dominic.dallosto@qut.edu.au}}). This work received funding from the Australian Government, via grant AUSMURIB000001 associated with ONR MURI grant N00014-19-1-2571. Authors acknowledge continued support from the Queensland University of Technology through the Centre for Robotics.
}%
}

\begin{document}

\AddToShipoutPicture*{%
     \AtTextUpperLeft{%
         \put(-3.5,10){
           \begin{minipage}{\textwidth}
              \scriptsize
              IEEE/RSJ International Conference on Intelligent Robots and Systems 2021\\
              Preprint version; final version available at \url{http://ieeexplore.ieee.org/}
           \end{minipage}}%
     }%
}

\bstctlcite{MyBSTcontrol}

\maketitle
\thispagestyle{empty}
\pagestyle{empty}
\input{tex/0-abstract}
\input{tex/1-introduction}
\input{tex/2-relatedworks}
\input{tex/3-methods}
\input{tex/4-results}
\input{tex/5-conclusion}

\noindent\textbf{Acknowledgements:} We would like to thank Suman Bista for kindly providing code for~\cite{bistaAppearancebasedIndoorNavigation2016a} for comparisons.
\bibliographystyle{IEEEtran}
\bibliography{references,references_ieeectrl}

\end{document}

%% file: math.tex
\newcommand{\tform}[2]{{}^{#1}\boldsymbol{T}_{#2}}
\newcommand{\tformInv}[2]{{}^{#1}\boldsymbol{T}^{-1}_{#2}}
\newcommand{\robotposeref}{\tform{T}{p}}

\newcommand{\robotposequery}{\tform{R}{\hat p}}
\newcommand{\robotposequeryInv}{\tformInv{R}{\hat p}}
\newcommand{\goalposeref}{\tform{T}{g}}
\newcommand{\goalposerefN}[1]{\tform{T}{g_{#1}}}

\newcommand{\goalposequery}{\tform{R}{\hat g}}
\newcommand{\goalposequeryN}[1]{\tform{R}{\hat{g}_{#1}}}
\newcommand{\goalposequeryNInv}[1]{\tformInv{R}{\hat{g}_{#1}}}
\newcommand{\goaloffsetimage}{\delta}

\newcommand{\goaloffsetrotationN}[1]{\theta_{\delta_{#1}}}
\newcommand{\rotation}{\boldsymbol{R}}
\newcommand{\corr}{\rho}
\newcommand{\corrthreshold}{\bar\corr}

\newcommand{\trans}[1]{\left({#1}\right)_{\boldsymbol{t}}}
\newcommand{\imageref}{\boldsymbol{I}}
\newcommand{\imagequery}{\boldsymbol{\hat I}}
\newcommand{\route}{\boldsymbol{R}}
\newcommand{\distancethreshold}{\tau_d}
\newcommand{\angularthreshold}{\tau_\alpha}
\newcommand{\nccsweepvar}{d}
\newcommand{\nccsweeprange}{D}
\newcommand{\alongpathsearchrange}{K}
\newcommand{\gainrotation}{K_\theta}
\newcommand{\gaindistance}{K_p}
\newcommand{\mean}[1]{\mu({#1})}
\newcommand{\FOV}{\rm{FOV}}
\newcommand{\imwidth}{I_w}
\newcommand{\imheight}{I_h}

\DeclareMathOperator*{\argmax}{arg\,max}

%% file: tex/0-abstract.tex
\begin{abstract}
Fully autonomous mobile robots have a multitude of potential applications, but guaranteeing robust navigation performance remains an open research problem. For many tasks such as repeated infrastructure inspection, item delivery, or inventory transport, a route repeating capability can be sufficient and offers potential practical advantages over a full navigation stack.
Previous teach and repeat research has achieved high performance in difficult conditions predominantly by using sophisticated, expensive sensors, and has often had high computational requirements.
Biological systems, such as small animals and insects like seeing ants, offer a proof of concept that robust and generalisable navigation can be achieved with extremely limited visual systems and computing power.
In this work we create a novel asynchronous formulation for teach and repeat navigation that fully utilises odometry information, paired with a correction signal driven by much more computationally lightweight visual processing than is typically required. This correction signal is also decoupled from the robot's motor control, allowing its rate to be modulated by the available computing capacity. We evaluate this approach with extensive experimentation on two different robotic platforms, the Consequential Robotics Miro and the Clearpath Jackal robots, across navigation trials totalling more than 6000 metres in a range of challenging indoor and outdoor environments. Our approach continues to succeed when multiple state-of-the-art systems fail due to low resolution images, unreliable odometry, or lighting change, while requiring significantly less compute. We also -- for the first time -- demonstrate versatile cross-platform teach and repeat without changing parameters, in which we learn to navigate a route with one robot and repeat that route using a completely different robot.
\end{abstract}

%% file: tex/1-introduction.tex
\section{Introduction}
\label{sec:introduction}

Visual navigation is a key capability for mobile robots to operate in a large range of environments without needing dedicated infrastructure -- vision being the primary sense humans use for navigating our world~\cite{ekstromWhyVisionImportant2015}. Navigation approaches are chiefly characterised by whether they use a map, and whether the map is metrically consistent or provides only topological information ~\cite{bonin-fontVisualNavigationMobile2008}. A general solution to the navigation problem remains elusive, but useful behaviours have been developed for limited ranges of conditions or environments.

Repeated route following is a useful capability for a mobile robot in many different applications: for a tour guide robot in a museum~\cite{burgardExperiencesInteractiveMuseum1999}, for a drone conducting environmental inspection~\cite{gallacherDroneApplicationsEnvironmental2017}, for automated mining transport trucks~\cite{robertsAutonomousControlUnderground2000}, or for an interplanetary rover repeatedly collecting samples. By considering that the robot only needs to traverse predetermined paths within the environment, the navigation problem can be solved by the teach and repeat framework~\cite{krajnikNavigationLocalisationReliable2018,warrenThereNoPlace2019,nguyenAppearanceBasedVisualTeachAndRepeatNavigation2016,furgaleVisualTeachRepeat2010}. Here, the robot is first manually driven along a route of interest while recording data. The recorded information is then used to robustly repeat this same route; however, the conditions between the teach and repeat runs might differ considerably in terms of lightning, moving obstacles, et cetera.

\begin{figure}[t]
    \vspace{0.2cm}
	\centering
	\includegraphics[width=0.95\linewidth]{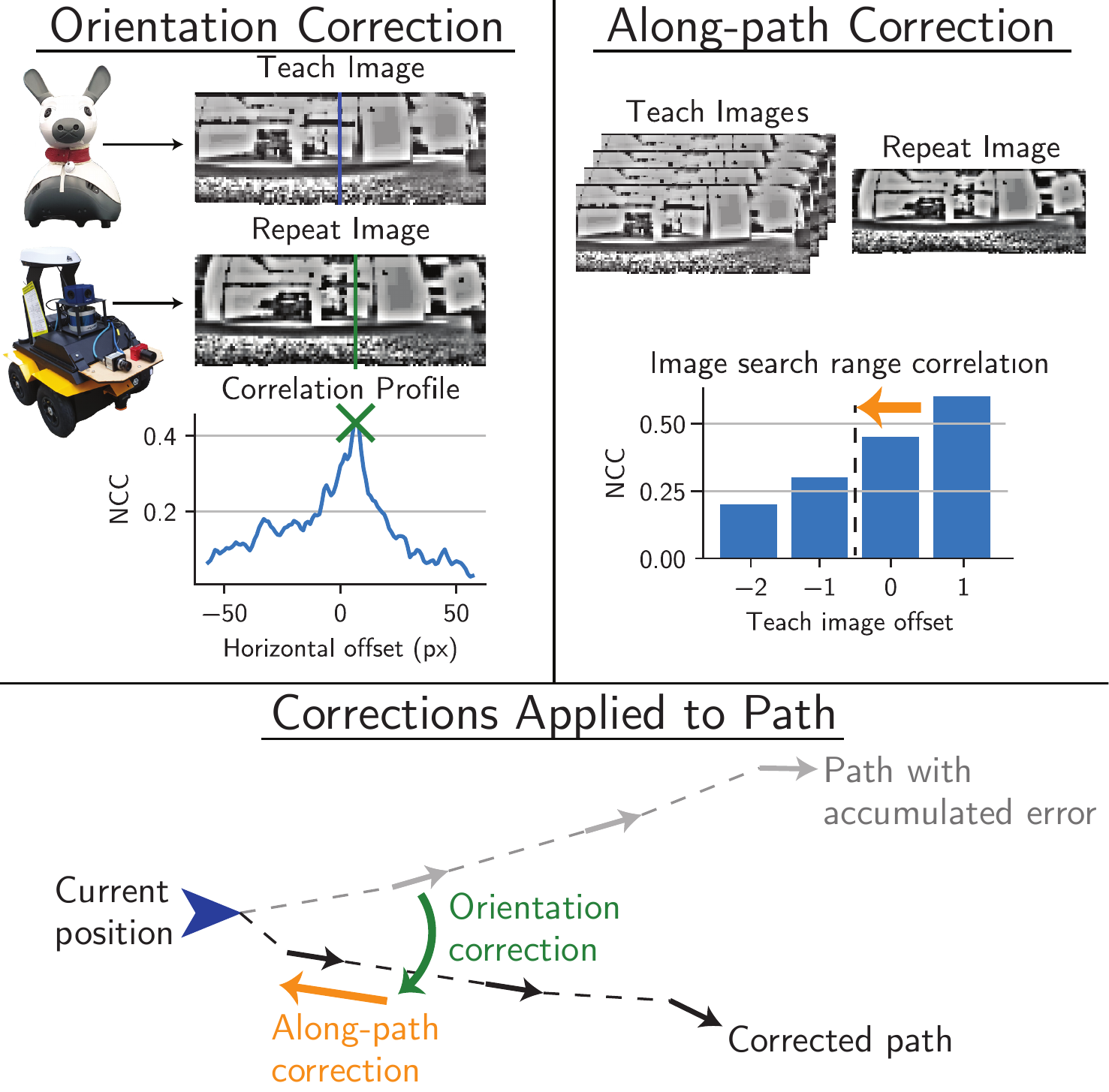}
	\vspace*{-0.2cm}
	\caption{Overview of our approach: Our teach and repeat system is predominantly driven by wheel odometry, with orientation and along-path corrections made using visual information. This framework enables fast and robust teach and repeat navigation, even in challenging situations like the teach and repeat runs being performed on very different robotic platforms (e.g.~Miro and Jackal).}
	\label{fig:overview}
	\vspace*{-0.15cm}
\end{figure}

Teach and repeat has been proven robust and flexible in many domains, from simple indoor environments~\cite{krajnikSimpleStableBearingonly2010,bistaAppearancebasedIndoorNavigation2016,bistaAppearancebasedIndoorNavigation2016a,bistaCombiningLineSegments2017,camaraAccurateRobustTeach2020} to the surface of another planet~\cite{furgaleVisualTeachRepeat2010,mcmanusVisualTeachRepeat2012,clementRobustMonocularVisual2017}. But even low compute teach repeat methods~\cite{krajnikSimpleStableBearingonly2010} still require high resolution images and significant computation for feature detection. This limits their deployment on small low-cost robots, for which a simple route following capability would be most useful. Such robots do not typically have stereo vision or LiDAR sensors. In contrast, almost all mobile robots have wheel odometry sensors and a monocular camera.

\vspace{0.05cm}
\noindent Our list of key contributions is the following:
\vspace{-0.075cm}
\begin{enumerate}[wide]
\item A novel, robust teach and repeat system that can be deployed on low-cost robots that have poor odometry and a low-resolution monocular camera. This approach performs robustly in difficult (low) lighting conditions and when occlusions occur, such as people in front of the robot.
\item Our system is predominantly driven by wheel odometry, so the complexity of the image processing needed to compute a periodic correction signal is significantly reduced. This allows the robot's basic navigation to be decoupled from the visual input, so that the correction rate can be changed as dictated by the available computation.
\item With over 6000 metres of navigation trials on two different robots in indoor and outdoor environments, we show that our approach is significantly more robust than the state-of-the-art in overall performance, as well as resilient to unreliable odometry and low resolution images.
\item Our approach can be deployed to new robots with minimal tuning. Indeed, our system is the first to demonstrate that the teach run from one robot can be repeated on a completely different robot.
\end{enumerate}

An overview of this teach and repeat system is shown in Fig.~\ref{fig:overview}. To foster future research, we make our code available for research purposes: \url{https://github.com/QVPR/teach-repeat/}.

%% file: tex/2-relatedworks.tex
\section{Related Works}
\label{sec:related-work}

Teach and repeat navigation falls under the broader capability of route following, which we briefly review in Section~\ref{subsec:routefollowing}. We then provide an overview of teach and repeat navigation approaches in Section~\ref{subsec:teachandrepeat}.

\subsection{Route Following}
\label{subsec:routefollowing}
Route following approaches for mobile robots are most easily distinguished by the types of map they use for navigation: metric; topological; or a hybrid of the two called topometric. We will introduce these three approaches in the following subsections, and provide some key references to relevant works.

\subsubsection{Metric Map}
A robot could explore and build an accurate metric map of the environment (as in Simultaneous Localisation and Mapping~\cite{cadenaPresentFutureSimultaneous2016}), and then use this map to track its location and repeat the desired route. A metric map allows for the most accurate route repetition, but maintaining a metrically consistent map scales in computational complexity with the size of the environment, so it is not commonly applicable to low-cost robots. Many robot navigation behaviours are still possible without a metric map~\cite{bonin-fontVisualNavigationMobile2008}. Indeed, it has recently been proven that local correction signals at each point along the path are sufficient for route following~\cite{krajnikNavigationLocalisationReliable2018}.

\subsubsection{Topological Map}
In visual servoing, a robot's goal is expressed as an image and the robot moves through the world to reach this location, without explicitly considering the world's geometry~\cite{chaumette2006visual}. Simple visual homing techniques are exhibited by ants and other insects to return to food or nesting sites~\cite{zeilVisualHomingInsects2009}. Visual homing can be extended for route following by using a topological map of the route -- considering the route as a sequence of visual goals and updating the current goal whenever the previous one is reached~\cite{courbonIndoorNavigationNonholonomic2008,bistaAppearancebasedIndoorNavigation2016a}.

However, visual homing techniques do not guarantee a smooth trajectory straight towards the goal~\cite{labrosseShortLongrangeVisual2007}. A sequential route representation is only minimally robust because failing to reach any goal causes the route following to fail completely. A purely visual servoing solution is therefore not well suited to precise robotic navigation.
Indeed, recent research debates whether ants combine path integration with vision to form a cognitive map, enabling more robust homing behaviours~\cite{webbInternalMapsInsects2019}.

\subsubsection{Topometric Map}
An alternative formulation is a topometric map~\cite{simhonGlobalTopologicalMap1998}, where locally accurate movement information such as wheel or visual odometry is combined with globally accurate sensor readings from a camera or LiDAR to form a map that is only \textit{locally} consistent. By not enforcing global consistency, topometric maps can efficiently scale to large environments such as in~\cite{zhangRobustAppearanceBased2009,furgaleVisualTeachRepeat2010}. Route following is then achieved through teach and repeat. In the teach portion the robot is driven once along the desired route, storing a series of sensory information. In the repeat run the robot then follows the trained route, robust to slight deviations from the path or variations in the environment.

\subsection{Teach and Repeat Navigation}
\label{subsec:teachandrepeat}
Early work on teach and repeat showed that following a route was possible by only making heading corrections along that route, without requiring a consistent map~\cite{matsumotoVisualNavigationUsing1996,burschkaVisionbasedControlMobile2001}. Further work developed a provably convergent control rule for this formulation~\cite{zhangRobustAppearanceBased2009}. Krajnik et al.~\cite{krajnikSimpleStableBearingonly2010} have proven teach repeat stable for a closed polygonal path, which was extended to arbitrary routes in work dubbed Bearnav~\cite{krajnikNavigationLocalisationReliable2018}.

Teach and repeat has also been employed in various domains: it has enabled autonomous navigation and emergency return capabilities for unmanned aerial vehicles in GPS denied environments~\cite{warrenThereNoPlace2019,nguyenAppearanceBasedVisualTeachAndRepeatNavigation2016}, and has also been used for autonomous underwater vehicles to repeat previous routes using sidescan sonar imagery~\cite{vandrishAUVRouteFollowing2012,kingPreliminaryFieldTrials2014}. In the following subsections we focus exclusively on teach and repeat for mobile robots, categorising approaches by the way they compare images between the teach and repeat runs.

\subsubsection{Direct visual methods}
The simplest approach is direct visual comparison, where full images are compared between the teach and repeat runs. These techniques can generally be run with lower resolution images so need the least computation, however they can be less robust to appearance change or large shifts in image location.

This can be ameliorated by image preprocessing techniques, and by assuming the robot's environment is flat so displacements will only occur horizontally in the image. For example,~\cite{matsumotoVisualNavigationUsing1996} used cross correlation to find the horizontal offset between teach and repeat images, providing the error signal for correction. 
Similarly, optimising the mutual information between reference and query images has been used to formulate a teach and repeat control rule that is robust to occlusion and lighting variations~\cite{dameUsingMutualInformation2013,bistaAppearancebasedIndoorNavigation2016}.

\subsubsection{Feature based methods}
In contrast to direct comparison methods, feature based techniques extract a number of salient features -- generally points, edges or corners -- from each image, and compare these between teach and repeat runs. Corresponding point locations can be used to estimate the geometrical offset between matched images. While typically more computationally intensive and with limitations in featureless environments, feature based comparison methods are the most widely used for teach and repeat. 
Early work derived a control scheme from the horizontal location of feature correspondences in monocular images~\cite{zhichaochenQualitativeVisionbasedMobile2006}, which was later made more robust by incorporating wheel odometry~\cite{zhichaochenQualitativeVisionBasedPath2009}.

To improve performance in challenging lighting conditions, depth images rendered from LiDAR scans were adopted in~\cite{mcmanusVisualTeachRepeat2012}. Furthermore, the approach was modified for monocular cameras in~\cite{clementRobustMonocularVisual2017}, where assuming a flat ground plane allowed recovery of depth information from single images. 

The feature based approach to teach and repeat approach has recently been extended with work from the visual place recognition field (see~\cite{lowryVisualPlaceRecognition2016} for a survey), where more detailed image descriptors are used to uniquely identify images in the world. For example, \cite{camaraAccurateRobustTeach2020} used image descriptors that are generated by a Convolutional Neural Network. This has the added benefit of being much more robust to appearance and viewpoint changes. Similarly, \cite{bistaAppearancebasedIndoorNavigation2016a} presented IBVS, a purely visual teach and repeat approach using image line features, as they are more robust to occlusion and motion blur. %

%% file: tex/3-methods.tex
\section{Proposed Approach}
\label{sec:methods}
This work presents a novel teach and repeat approach, reformulating the provably convergent correction rule from ~\cite{zhangRobustAppearanceBased2009} into a geometric framework, where each waypoint in the route is represented as a pose relative to the robot's odometry frame. This geometric formulation allows odometry information from the reference run to be more efficiently used when the route is repeated. Additionally, this formulation decouples odometry information from the visual information stream so that the robot can be smoothly controlled at high frequencies with less frequent image corrections, allowing for reduced computation and graceful performance degradation.
Here we detail the approaches to the two distinct phases of operation for the robot: teach and repeat.

The following notation is used, aiming to be consistent with~\cite{corkeRoboticsVisionControl2011}: $\tform{A}{B}$ represents the pose of $B$ relative to $A$, and is a homogeneous transformation matrix, $\tform{}{} \in \mathit{SE} (2) \subset \mathbb{R}^{3\times 3}$.

\subsection{Teach phase}
\label{subsec:teach}
In the teach phase the robot is manually teleoperated along the desired route, while recording a list of associated odometry poses and images. As the odometry information suffers from drift and is not globally consistent, this results in a topometric map of the route. This topometric map, $\route$, is an ordered list of cardinality $N$, containing pairs of 3~degrees-of-freedom odometry poses, $\goalposeref$, and images, $\imageref$.
\begin{align}
	\route &= \{(\goalposerefN{1}, \imageref_{1}), (\goalposerefN{2}, \imageref_{2}), \dots, (\goalposerefN{N}, \imageref_{N})\}
\end{align}
A new entry $(\goalposerefN{N+1}, \imageref_{N+1})$ is appended to the route whenever the displacement of the robot from the last recorded pose exceeds a certain distance $\distancethreshold$ or angular threshold $\angularthreshold$. Lower thresholds allowed for denser maps and more accurate path following at the expense of a greater memory requirement.\looseness=-1

Before being saved, the images are preprocessed by downscaling them, converting to greyscale, and applying patch normalisation. Downscaling significantly reduces the memory required to store the route, and patch normalisation~\cite{milfordSeqSLAMVisualRoutebased2012} affords the system some robustness to lighting variation.

\subsection{Repeat phase}
In the repeat phase, the robot tries to best follow the route stored in its topometric map. Control is structured hierarchically, with a low level odometry-based controller driving the robot to a nearby pose in its odometry frame, $\goalposequery$ (see Section~\ref{subsub:odometrycontrol}), and a route correction controller updating the target pose based on the current image \mbox{frame, $\imagequery$}. These corrections are computed separately for orientation -- to account for heading and lateral path errors (Section~\ref{subsub:orientationcorrection}) -- and along-path errors (Section~\ref{subsub:alongpathcorrection}).

\subsubsection{Odometry driven control}
\label{subsub:odometrycontrol}
The controller receives a high frequency odometry-based estimate of the robot's pose, $\robotposequery$, and drives the robot's motors so it reaches a goal pose, $\goalposequery$. Both of these poses are expressed in the robot's local odometry frame. The pose controller detailed in~\cite[Sec. 4.2.4]{corkeRoboticsVisionControl2011} was chosen for this implementation, but in principle any equivalent controller could be substituted.

\subsubsection{Image based orientation correction}
\label{subsub:orientationcorrection}
Incoming images, $\imagequery$, are preprocessed as described in Section~\ref{subsec:teach}. Normalised Cross Correlation~(NCC)~\cite{lewisFastNormalizedCrossCorrelation2001} is used to compare the incoming images to those saved during the teach run, $\imageref$. Assuming that the robot operates in a flat planar world, rotational and lateral offsets from the path both cause horizontal image displacements. For example, a left path offset and an anticlockwise rotation both cause an image offset to the right, so cannot be distinguished. But because both can be corrected by turning clockwise, the ambiguity between rotational and lateral offsets is acceptable. Specifically, the query image is swept horizontally over a search range, $\nccsweepvar \in [-\nccsweeprange,\nccsweeprange]$ (see Table~\ref{tab:params}), and the NCC computed with the current reference image for each of these offsets: 
\begin{align}
	\label{eq:NCC}
		&\text{NCC}_\nccsweepvar(\imageref_n,\imagequery) = \\
		&\frac{\sum\limits_{x,y} \left[
		\big( \imageref_n(x,y) -  \mean{\imageref_{n,{\nccsweepvar}}} \big)
		\big( \imagequery(x-\nccsweepvar,y) - \mean{\imagequery} \big)\right]}
		{\left(
		\sum\limits_{x,y}\Big[\imageref_n(x,y) - \mean{\imageref_{n,{\nccsweepvar}}} \Big]^2 
		\sum\limits_{x,y}\Big[\imagequery(x-\nccsweepvar,y) - \mean{\imagequery} \Big]^2
		\right)^{0.5}},\nonumber
\end{align}
where $n$ is the index of the current reference image in the teach run, $x$ and $y$ are incremented through the range of valid reference image pixel coordinates overlapping the shifted query image: $x \in [\max(0,\nccsweepvar),\min(\imwidth,\imwidth+\nccsweepvar))$ and $y \in [0,\imheight)$, $\imwidth$ and $\imheight$ are the width and height of the images respectively, $\mean{\imagequery}$ is the mean of the query image, and $\mean{\imageref_{n,{\nccsweepvar}}}$ is the mean of the region of the reference image that overlaps with the shifted query image. This results in a correlation profile as in Fig.~\ref{fig:overview}. The offset with the greatest NCC, 
\begin{align}
	\goaloffsetimage_n = \argmax_{\nccsweepvar \in [-\nccsweeprange,\nccsweeprange]} \left(\text{NCC}_\nccsweepvar(\imageref_n,\imagequery)\right), 
\end{align}
is used as the offset estimate. This pixel offset is converted to an offset angle as if the offset were purely rotational: $\goaloffsetrotationN{n} = \frac{\FOV}{\imwidth} \goaloffsetimage_n$, where $\FOV$ is the horizontal angular field of view of the image.

\begin{figure}[t]
	\centering
	\vspace{0.2cm} %
	\includegraphics[scale=0.7]{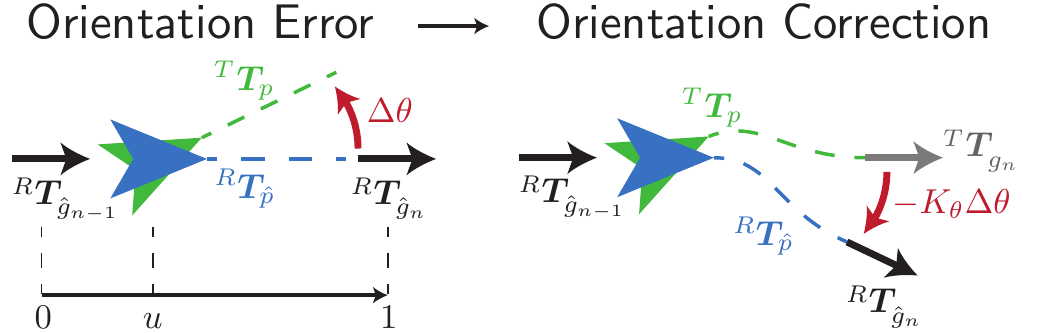}
	\caption{Orientation correction: An error between the robot's estimated pose, $\robotposequery$, and actual pose, $\robotposeref$, is calculated by interpolating the image offsets between the previous goal, $\goalposequeryN{n-1}$, and current goal, $\goalposequeryN{n}$, with interpolation factor, $u$. This error, $\Delta\theta$, is multiplied by the gain factor $\gainrotation$, to rotate the target path so the robot correctly follows the taught route.}
	\label{fig:orientation-localisation}
\end{figure}

Following~\cite{zhangRobustAppearanceBased2009}, image offsets are interpolated between the previous and next goals to allow for corrections to be made between goal locations. Being in front of or behind a goal could induce a horizontal image offset in either direction, depending on the dominant visual features in the scene. An object on the left will appear to move further left if the robot advances, while an object on the right will appear to move right. Interpolating the offsets between goals cancels out this along-path induced offset, allowing it to be distinguished from rotational or lateral path offsets, which affect both $\goaloffsetimage_{n-1}$ and $\goaloffsetimage_n$ similarly.

The interpolation factor, $u$, is the proportion of the distance travelled between two keyframe poses. Specifically, $u$ can be calculated with odometry information as follows:
\begin{align}
	\label{eq:u}
	u=\frac{\trans{\goalposequeryNInv{n-1} \goalposequeryN{n}} \cdot \trans{\goalposequeryNInv{n-1} \robotposequery}}
	{\left|\left|\trans{\goalposequeryNInv{n-1} \goalposequeryN{n}}\right|\right|^2},
\end{align}
where $\robotposequery$ is the robot's pose estimate during the repeat run, and $\goalposequeryN{n}$ is the pose of goal $n$ in the robot's odometry frame (see Fig.~\ref{fig:orientation-localisation}), $\trans{\tform{}{}}=(\tform{}{i,j})_{1\leq i \leq 2;\ j=3}$ extracts the translation component of the pose $\tform{}{}$, $\cdot$ is the dot product of two vectors, and $||\boldsymbol{t}||$ denotes the magnitude of vector, $\boldsymbol{t}$.

The orientation offset of the robot from the path at this point between the two keyframes is interpolated as follows:
\begin{align}
	\Delta\theta &= (1-u)\goaloffsetrotationN{n-1} + u\goaloffsetrotationN{n},
\end{align}
where $\goaloffsetrotationN{n}$ is the rotational offset to keyframe $n$. To correct this error, the current goal pose is rotated about the robot in the opposite direction to the visual offset:
\begin{align}
	\goalposequeryN{n} &\leftarrow \robotposequery \left(\rotation({-\gainrotation\Delta\theta}) \robotposequeryInv \goalposequeryN{n}\right),
\end{align}
where $\rotation(\theta)$ is a $3\times 3$ homogeneous transformation matrix causing a rotation of \mbox{angle $\theta$}, and $\gainrotation$ is a calibrated gain term (gain parameter selection is discussed in Section~\ref{sec:sensitivity}).

\subsubsection{Image based along-path correction}
\label{subsub:alongpathcorrection}

In addition to lateral or rotational errors, the robot can have an offset from travelling too quickly or slowly along the path. For this case, the assumption was made that the current image would correlate most strongly with goal images closest to the robot's current location. A small search range, $\pm \alongpathsearchrange$, is centred on the previous and current goal images, i.e. $\imageref_{n-2}$ to $\imageref_{n+1}$ for $\alongpathsearchrange = 1$, and correlation values compared for this range. Higher correlation values ahead of the robot pull it forwards along the path, and vice versa. This provides incremental corrections to account for random and systematic odometry errors. For goals at the start or end of the route, the search range is symmetrically scaled down to be completely valid.

\begin{figure}[t]
    \vspace{0.2cm} %
	\centering
	\includegraphics[scale=0.7]{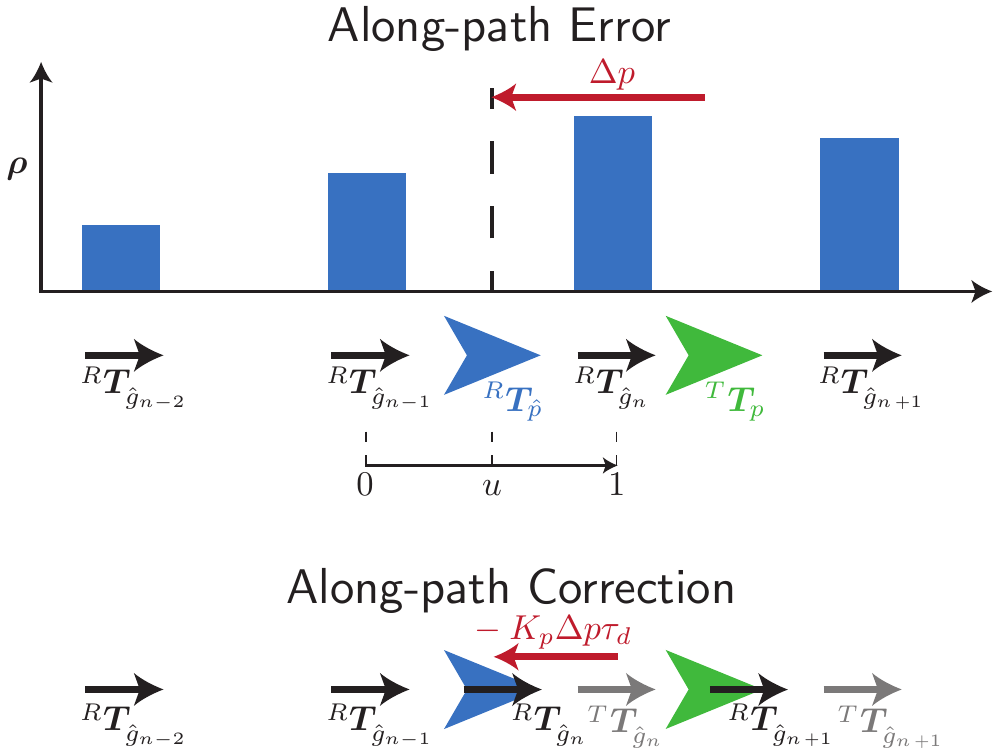}
	\caption{Along-path correction: An error between the robot's estimated pose, $\robotposequery$, and actual pose, $\robotposeref$, is calculated by finding the goal, $\goalposequery$, with the image that correlates most strongly, $\rho$, to the current image, relative to $u$. This error, $\Delta p$, is multiplied by a gain factor, $\gaindistance$, converted to a scaling parameter, $s$, and used to extend or retract the target path so the robot correctly follows the taught route.}
	\vspace{-0.25cm}
	\label{fig:alongpath-localisation}
\end{figure}

Peak correlations between the query and reference images over the search range are rectified with respect to a threshold to remove the effects of noise-level correlations: 
\begin{gather}
	\boldsymbol{\corr} = \left\{ \max\left(0, \max_{\nccsweepvar \in [-\nccsweeprange,\nccsweeprange]} \left(\text{NCC}_\nccsweepvar(\imageref_{n+k},\imagequery)\right) - \corrthreshold\right)\right\}_{k=-(\alongpathsearchrange+1)}^\alongpathsearchrange
	\raisetag{20pt}
\end{gather}
where $\corrthreshold$ is the noise-level correlation threshold (see Table~\ref{tab:params}) and $\text{NCC}_\nccsweepvar$ is defined in Eq.~(\ref{eq:NCC}). The weighted average of the rectified correlations is then taken to compute an estimate of the robot's along path error, $\Delta p$, as follows:
\begin{align}
	\Delta p &= \frac{\sum\limits_{k=-(\alongpathsearchrange+1)}^\alongpathsearchrange\left( k\corr_k \right)}
	{\sum\limits_{k=-(\alongpathsearchrange+1)}^\alongpathsearchrange\left( \corr_k \right)} - u,
\end{align}
where $u$ is the proportion of the distance travelled from the previous to the current goal, defined in Eq.~(\ref{eq:u}). This calculation is depicted in Fig.~\ref{fig:alongpath-localisation}. The position estimate is in units of goals (spaced $\distancethreshold$ apart), so is converted to a linear scaling correction factor, $s$, which is used to move the current goal towards or away from the robot:
\begin{align}
	s \crel{=} \frac{\left|\left|\trans{\robotposequeryInv \goalposequeryN{n}}\right|\right| - \gaindistance\Delta p \distancethreshold}{\left|\left|\trans{\robotposequeryInv \goalposequeryN{n}}\right|\right|} \\
	\trans{\goalposequeryN{n}} \crel{\leftarrow} s \trans{\goalposequeryN{n}},
\end{align}
where $\gaindistance$ is a calibrated gain term, and $\distancethreshold$ is defined in Section~\ref{subsec:teach}. 

If the correlation values are higher for images behind the robot's estimated location, i.e.~$\Delta p < 0$, such that $s > 1$, the current goal will be extended, giving the true pose, $\robotposeref$, time to ``catch up'' to the estimated pose, $\robotposequery$. The opposite case occurs if the robot estimates its position to be behind its true pose. Fig.~\ref{fig:alongpath-localisation} visualises the along-path correction.

\subsubsection{Global Initialisation}
When starting to repeat, the NCC (Eq.~\ref{eq:NCC}) is computed for all images in the route and the best match selected as the starting point. This allows the robot to localise to within $\frac{\distancethreshold}{2}$ of its true pose, well within the along-route search range, $(2\alongpathsearchrange+1)\distancethreshold$. If the NCC is below a threshold, the robot can detect that it is not located along the route and so does not start route repetition. An operator would need to manually reposition the robot closer to the route before route following can begin.

%% file: tex/4-results.tex
\section{Experimental Results}
\label{sec:results}
This section presents the results of over 6000 metres of trials, both indoor and outdoor, comparing our approach to two state-of-the-art benchmark systems~\cite{bistaAppearancebasedIndoorNavigation2016a,krajnikNavigationLocalisationReliable2018}. We first outline the experimental procedure and the two robots used in Section~\ref{subsec:experimentalprocedure}. This is followed by a set of comprehensive indoor trials in Section~\ref{subsec:indoortrials}, showing that our approach is robust to odometry errors, works with images of a very low resolution, and is not sensitive to calibrated parameter values. In Section~\ref{subsec:outdoortrials} we demonstrate that our approach can be applied outdoors on a 550~m route, at different times of the day and in varying weather conditions. 4 repeat runs were performed one month after the teach run, and one repeat run four months after the teach run. Both current state-of-the-art systems fail in these challenging conditions. Finally, \mbox{Section~\ref{subsec:transferteach}} demonstrates cross-platform versatility, where a teach run was recorded on the Miro robot and then repeated on the Jackal, the two robots having significant differences in their cameras, viewpoints, and odometry accuracies.

\begin{figure}[t]
	\centering
	\vspace{0.2cm} %
	\includegraphics[width=0.74\linewidth]{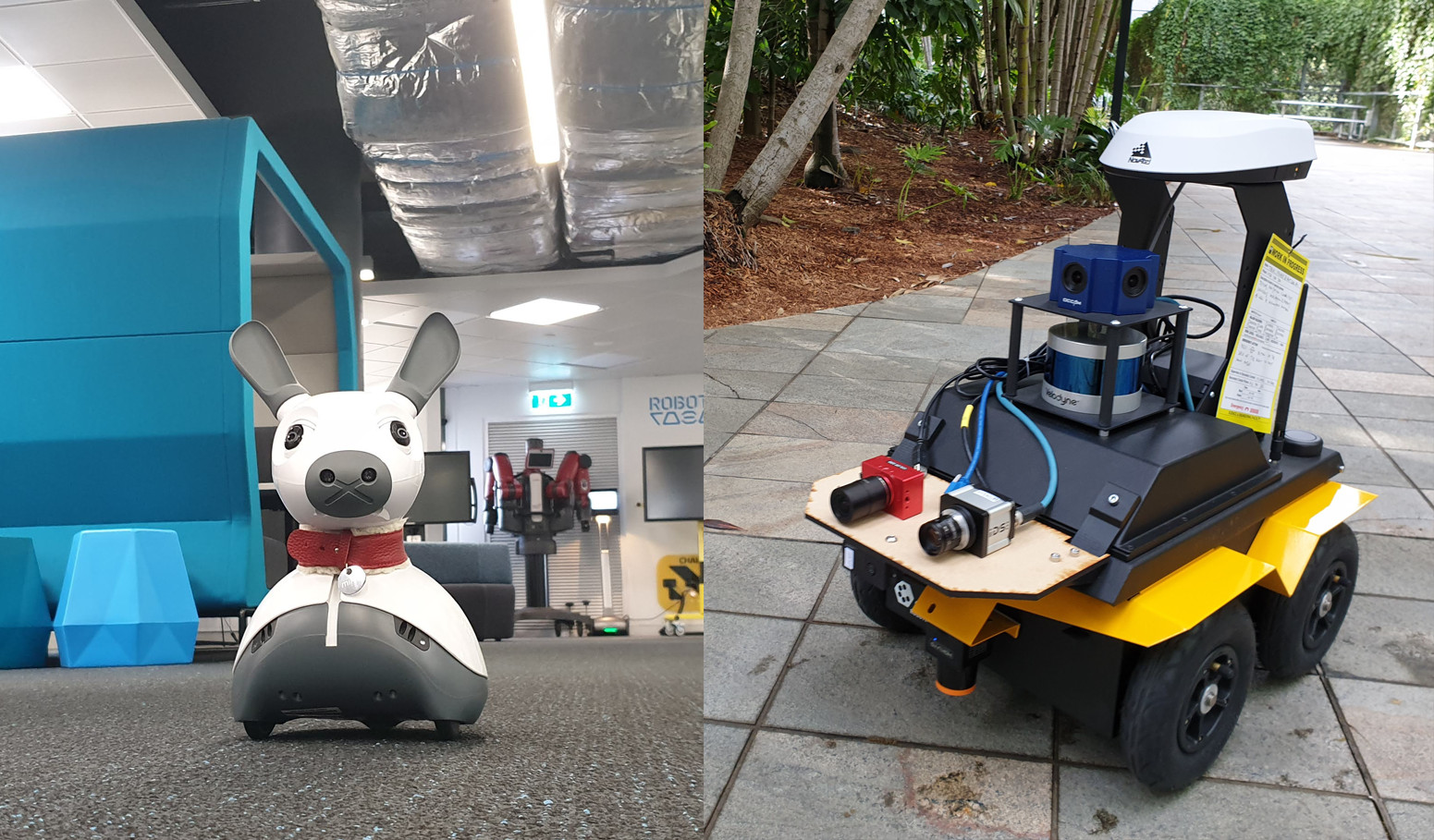}\\
	\scriptsize Miro view\\[0.05cm]
	\includegraphics[width=0.74\linewidth,trim=0 6.75cm 0 0.5cm, clip]{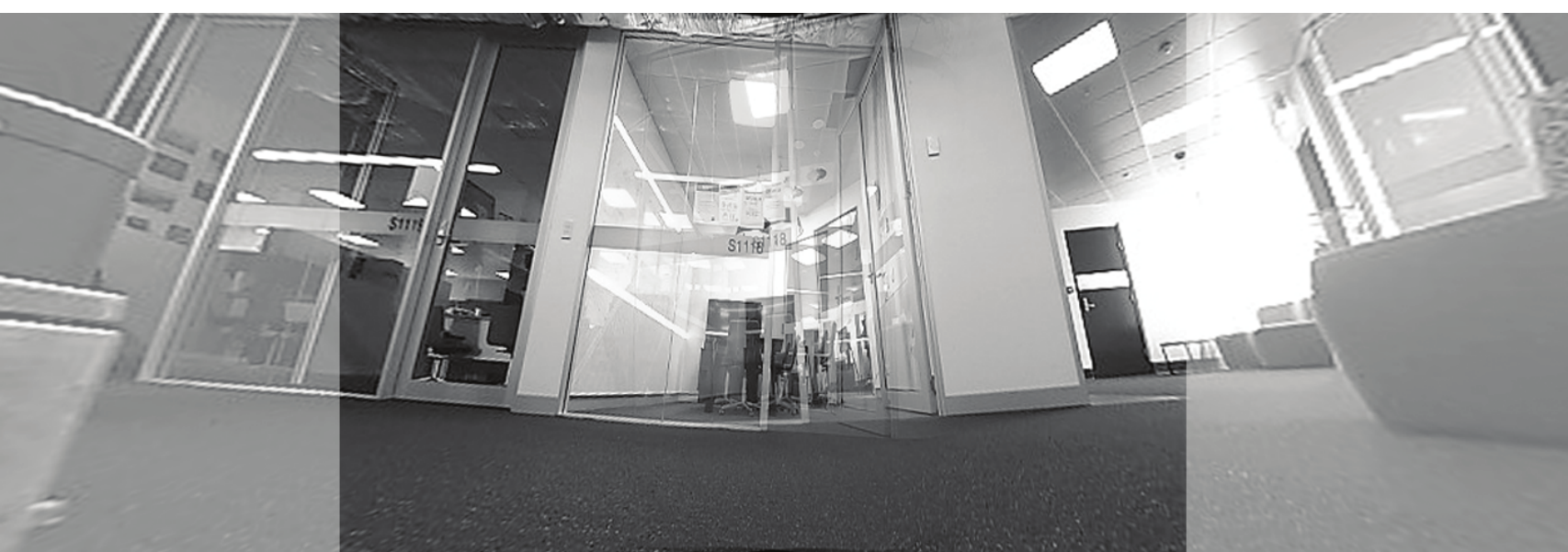}\\
	\scriptsize Jackal view\\[0.05cm]
	\includegraphics[width=0.74\linewidth,trim=0 0 0 7.25cm, clip]{figures/jackal-vs-miro-view.pdf}
	\vspace*{-0.15cm}
	\caption{(Top) Robotic platforms considered in this paper: Miro traversing the indoor route (left) and Jackal traversing the outdoor route (right). The front facing camera from the blue omnidirectional camera was used on the Jackal. For the Miro, images from the two eyes were stitched together into one with some distortion artefacts. (Bottom) Example views from Miro and Jackal. Miro's view was cropped to approximately match that of Jackal when transferring the teach run between robots as described in Section~\ref{subsec:transferteach}.}
	\vspace*{-0.1cm}
	\label{fig:robots}
\end{figure}

\subsection{Experimental Procedure}
\label{subsec:experimentalprocedure}
The proposed teach repeat system was run on the Jackal and Miro robots (Fig.~\ref{fig:robots}) in both indoor and outdoor environments (also see our Multimedia Material). All robot parameters were set as shown in Table~\ref{tab:params} unless otherwise specified. We note that the same parameters were used indoors and outdoors, and only a single parameter ($\tau_d$) differed between the two robot platforms (however $\tau_d$ was set to the same value on both platforms in Section~\ref{subsec:transferteach} without any loss in performance). 

We ran SLAM~\cite{Macenski2019} using the Jackal's Velodyne LiDAR sensor to measure how closely the repeat run matched the teach run. This SLAM process was run entirely independently from our teach and repeat approach -- no information was shared between them. Although not real ground truth, SLAM was deemed to be suitably accurate for metric error analysis indoors. Large relocalisation jumps occurred in the outdoor trials, so these data are only used for visualisation.

Additionally, we compare to two state-of-the-art benchmark systems, Bearnav~\cite{krajnikNavigationLocalisationReliable2018} and IBVS~\cite{bistaAppearancebasedIndoorNavigation2016a}. These two systems are, to the best of our knowledge, the only open-source teach and repeat methods. Bearnav has recently been used as a comparison method in~\cite{camaraAccurateRobustTeach2020}.

\begin{table}[!t]
    \vspace{0.2cm} %
    \centering
	\footnotesize
    \caption{Parameter values for the two robotic platforms}
    \vspace*{-0.1cm}
	\label{tab:params}
    \renewcommand{\arraystretch}{1.1}
    \begin{tabular}{c|c|c}
        Parameter & Jackal value & Miro value \\
        \hline
        Image size, $\imwidth\times\imheight$ & $115\times44$ & $115\times44$ \\
        Patch normalisation size & $9\times9$ & $9\times9$ \\
        NCC search range, $D$ (px) & $\pm 75$ & $\pm 75$ \\
        Noise correlation threshold, $\corrthreshold$ & $0.1$ & $0.1$ \\
		Horizontal field of view, $\FOV$ ($^\circ$) & $75$ & $175.2$ \\
        \hline
        Distance threshold, $\distancethreshold$ (m) & $0.3$ & $0.2$ \\
        Angle threshold, $\angularthreshold$ ($^\circ$) & $15$ & $15$ \\
        \hline
        Orientation correction gain, $\gainrotation$ & $0.01$ & $0.01$ \\
		Along-path correction gain, $\gaindistance$ & $0.01$ & $0.01$ \\
		Along-path search range, $\alongpathsearchrange$ & $3$ & $3$ \\
    \end{tabular}
    \vspace*{-0.25cm}
\end{table}

\subsection{Indoor Comprehensive Trials}
\label{subsec:indoortrials}
These experiments were performed on the Jackal robot in an office environment for which a high resolution SLAM map was already available, enabling accurate characterisation of the system's performance. Again, we emphasise that SLAM was run completely independently from the teach and repeat navigation, and was only used to measure performance. In particular we examine the robustness of the system to inaccurate odometry and low resolution images, and its sensitivity to parameter tuning.

\subsubsection{Robustness to Odometry Errors}

Robustness to inaccurate odometry information was determined to be a key performance characteristic of the system, particularly for application to low cost robots for which no other self motion sensor information, such as an IMU, is available. Significant odometry variations were for example observed when operating robots on different floor surfaces, or as the tyre pressure dropped. As in~\cite{zhangRobustAppearanceBased2009,krajnikNavigationLocalisationReliable2018}, orientation errors in odometry were found to be less critical, so here we examine the effects of an up to $\pm30\%$ systematic artificial corruption of linear odometry measurements. Teach run odometry information was kept unchanged for all trials.

\begin{figure}[t]
    \vspace{0.2cm} %
	\centering
    \includegraphics[width=0.98\linewidth]{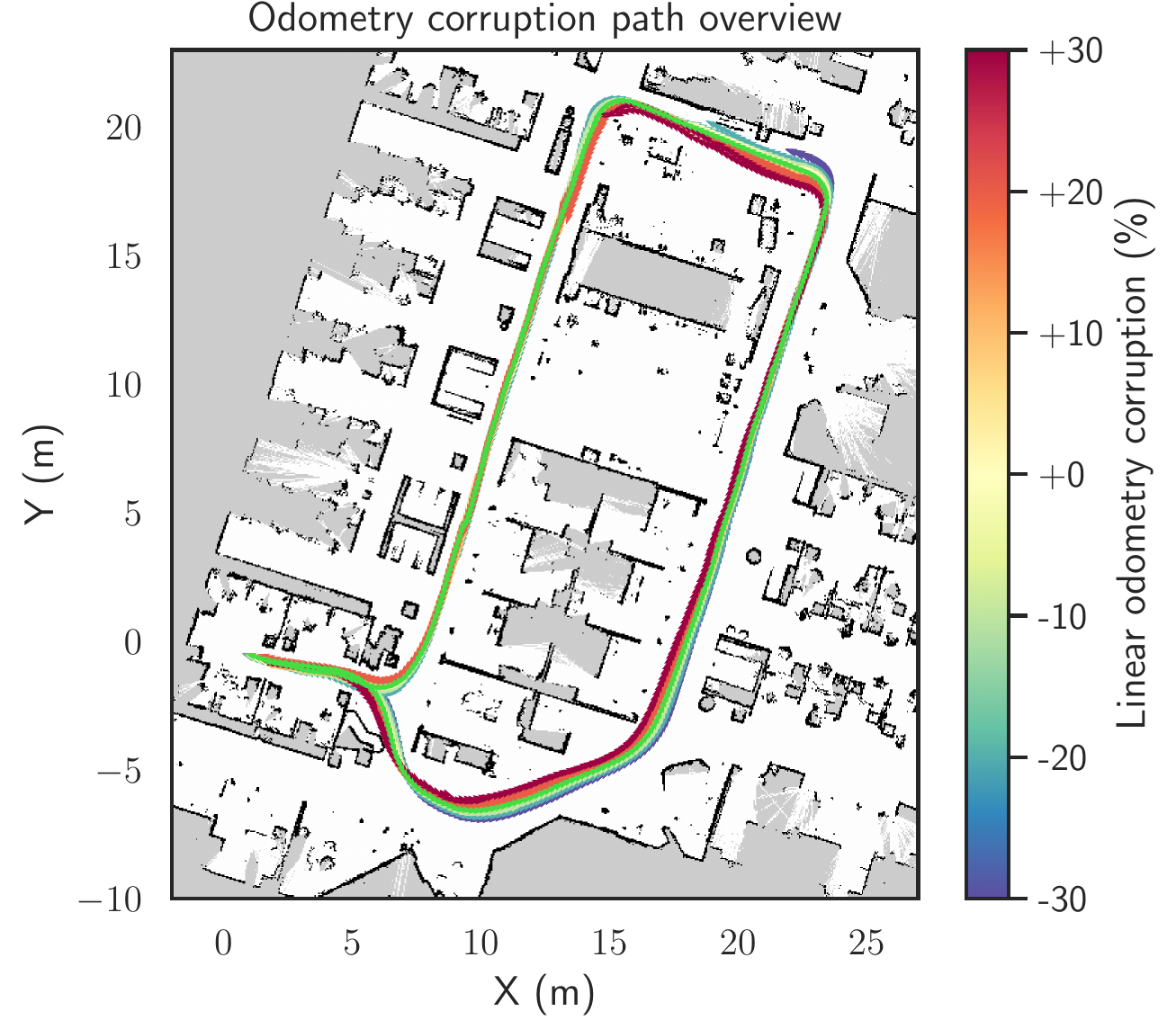}
    \vspace*{-0.4cm}
	\caption{Overview of the system's indoor traverses with corrupted odometry: One can observe increasing deviations from the teach run (green line) with increasing odometry corruption multiplier. Please refer to Fig.~\ref{fig:odom-analysis} for a quantitative analysis of these results.}
	\vspace{-0.4cm}
	\label{fig:indoor-runs}
\end{figure}

\begin{figure*}[t]
    \vspace{0.2cm} %
	\centering
	\includegraphics[width=0.8\linewidth]{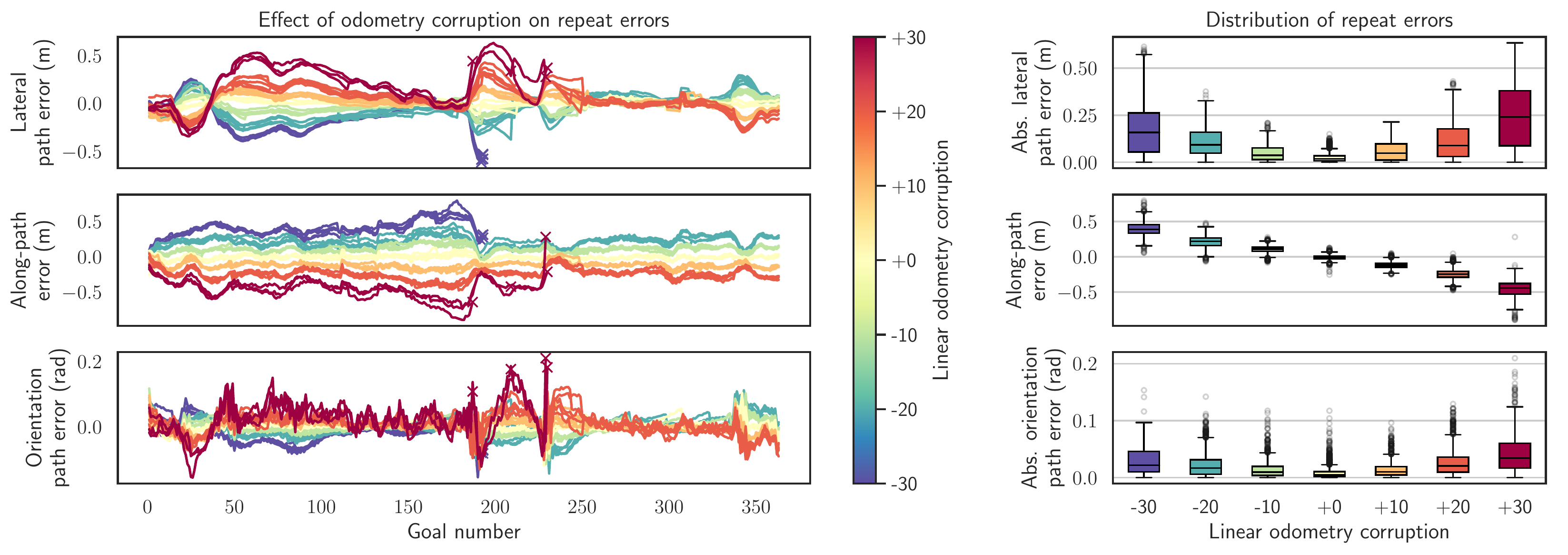}
	\vspace*{-0.25cm}
	\caption{System robustness to corrupted linear odometry. Errors grow smoothly with increased linear odometry corruption. Along-path errors are as expected for positive or negative odometry errors. Crosses indicate failure -- when manual intervention was required to avoid a crash. %
	}
	\vspace*{-0.45cm}
	\label{fig:odom-analysis}
\end{figure*}

The results for 5 trials at each of the odometry corruption values are shown in Figs.~\ref{fig:indoor-runs} and \ref{fig:odom-analysis}, and summarised in Table~\ref{tab:odomcorruption}. Errors during the repeat runs increased gradually with increasing odometry errors so performance remained robust for the range of $\pm 20\%$ corruption. Failures occurred for $\pm 30\%$ odometry corruption, when lateral path error exceeded the width of the corridors in which the robot was travelling and manual intervention was required to avoid crashing. This result is in comparison to a similar teach and repeat approach in~\cite{zhangRobustAppearanceBased2009}, which was only successful for $-5\%$ to $+10\%$ odometry error following a much simpler indoor route.

Additionally, we tested the Bearnav system~\cite{krajnikNavigationLocalisationReliable2018} in the same conditions. It explicitly does not account for errors in odometry distance measurements, so served as a good benchmark to test whether our along-path correction improved robustness. Indeed, Bearnav was successfully able to repeat the route with accurate odometry but failed consistently for even $\pm 5\%$ odometry error. IBVS~\cite{bistaAppearancebasedIndoorNavigation2016a} does not use odometry, relying solely on visual information, but it was still able to successfully repeat the indoor route.

\begin{table}[!t]
    \vspace{0.2cm} %
    \centering
    \setlength\tabcolsep{0.1cm}
    \footnotesize
    \caption{Success rate for different odometry corruption values}
    \vspace*{-0.1cm}
    \label{tab:odomcorruption}
    \begin{tabular}{c|c|c|c|c|c|c|c}
        Odometry\Tstrut\Bstrut & -30\% & -20\% & -10\% & +0\% & +10\% & +20\% & +30\% \\
        \hline
        Success ours\Tstrut\Bstrut & 0/5 & 5/5 & 5/5 & 5/5 & 5/5 & 5/5 & 0/5 \\
        Success Bearnav~\cite{krajnikNavigationLocalisationReliable2018} & 0/5 & 0/5 & 0/5 & 5/5 & 0/5 & 0/5 & 0/5\\
        Success IBVS~\cite{bistaAppearancebasedIndoorNavigation2016a} & -- & -- & -- & 5/5 & -- & -- & --\\
    \end{tabular}
\end{table}

\begin{table}[!t]
    \centering
    \footnotesize
    \caption{Success rate for different correction gains}
    \vspace*{-0.1cm}
    \label{tab:sensitivity}
    \begin{tabular}{c|c|c|c|c|c}
        Orient. corr. gain\Tstrut\Bstrut\ $\gainrotation$\Tstrut\Bstrut & 0 & 0.0001 & 0.001 & 0.01 & 0.1 \\
        \hline
        Success\Tstrut\Bstrut & 0/3 & 0/3 & 3/3 & 3/3 & 2/3 \\\multicolumn{6}{c}{}\\
        Along-path corr. gain\Tstrut\Bstrut\ $\gaindistance$ & 0 & 0.0001 & 0.001 & 0.01 & 0.1 \\
        \hline
        Success\Tstrut\Bstrut & 3/3 & 3/3 & 3/3 & 3/3 & 0/3
    \end{tabular}
    \vspace*{-0.1cm}
\end{table}

\subsubsection{Robustness to Reduced Image Size}
\label{subsec:imageresolution}
\begin{figure}[t]
	\centering
	\includegraphics[width=0.8\linewidth]{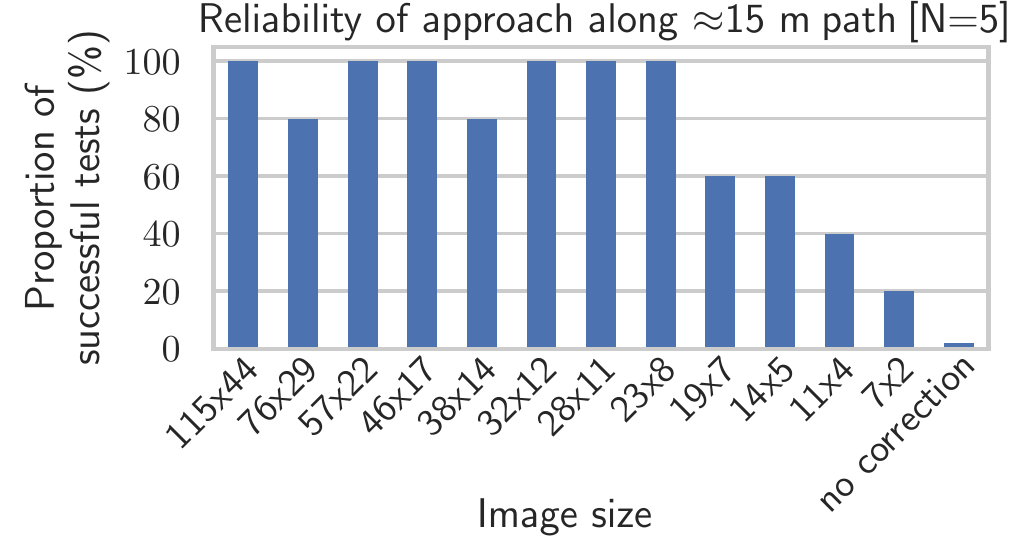}
	\vspace*{-0.3cm}
	\caption{System robustness to low resolution images, running on Miro robot. Performance only drops significantly for image resolutions lower than $23\times8$, with occasional failures at higher resolutions.}
	\label{fig:resolution}
	\vspace*{-0.5cm}
\end{figure}

Motivated by the successful navigation strategies of ants with their limited visual systems~\cite{zeilVisualHomingInsects2009}, we tested the system's robustness to low resolution images. Fig.~\ref{fig:resolution} shows results for Miro repeating a short (${\approx}15$~m) but challenging outdoor route. Our approach performs reliably with image resolutions as low as $23\times8$ before performance degrades significantly.

On the Jackal robot, we achieved reliable performance with $32\times12$ images indoors, and $57\times22$ images outdoors (teach in bright sunlight, repeat overcast). We generally observed that lower resolution images were sufficient when visual conditions were similar for teach and repeat, but higher resolution images were required when conditions changed.

This compares to Bearnav~\cite{krajnikNavigationLocalisationReliable2018}, which failed for resolutions below $140\times90$ in the indoor setting, due to a failure of image feature detection. Similarly, IBVS~\cite{bistaAppearancebasedIndoorNavigation2016a} failed for resolutions below $188\times120$. As expected, direct visual comparison techniques like ours proved more robust to low resolution images than feature based techniques, allowing use of 30 times smaller images compared to the state-of-the-art.

\subsubsection{Sensitivity Analysis of Teach Repeat Parameters}
\label{sec:sensitivity}

A sweep of the correction gain parameters was performed to determine the sensitivity of the system to calibration. The default value was empirically chosen to be 0.01 for both corrections, and only one parameter was varied at a time. The results in Table~\ref{tab:sensitivity} show the system working over at least one order of magnitude for both parameters. 

The system still worked without any path correction, but this was expected when compared to the successful repeat results for Bearnav, which does not implement path correction. Over the distance of the indoor run, the Jackal's linear odometry accuracy was sufficient for path following.

To stress the importance of the path correction, we conducted additional experiments where we set $K_p=0$ and artificially corrupted the linear odometry as in the previous experiment. In this case, the repeat runs failed consistently even for a relatively moderate corruption of $\pm10\%$, confirming the utility of the along-path correction introduced in Section~\ref{subsub:alongpathcorrection}.

\subsection{Extension Test 1: Long Distance Outdoor Trials}
\label{subsec:outdoortrials}
\begin{figure}[t]
	\centering
	\includegraphics[width=0.91\linewidth]{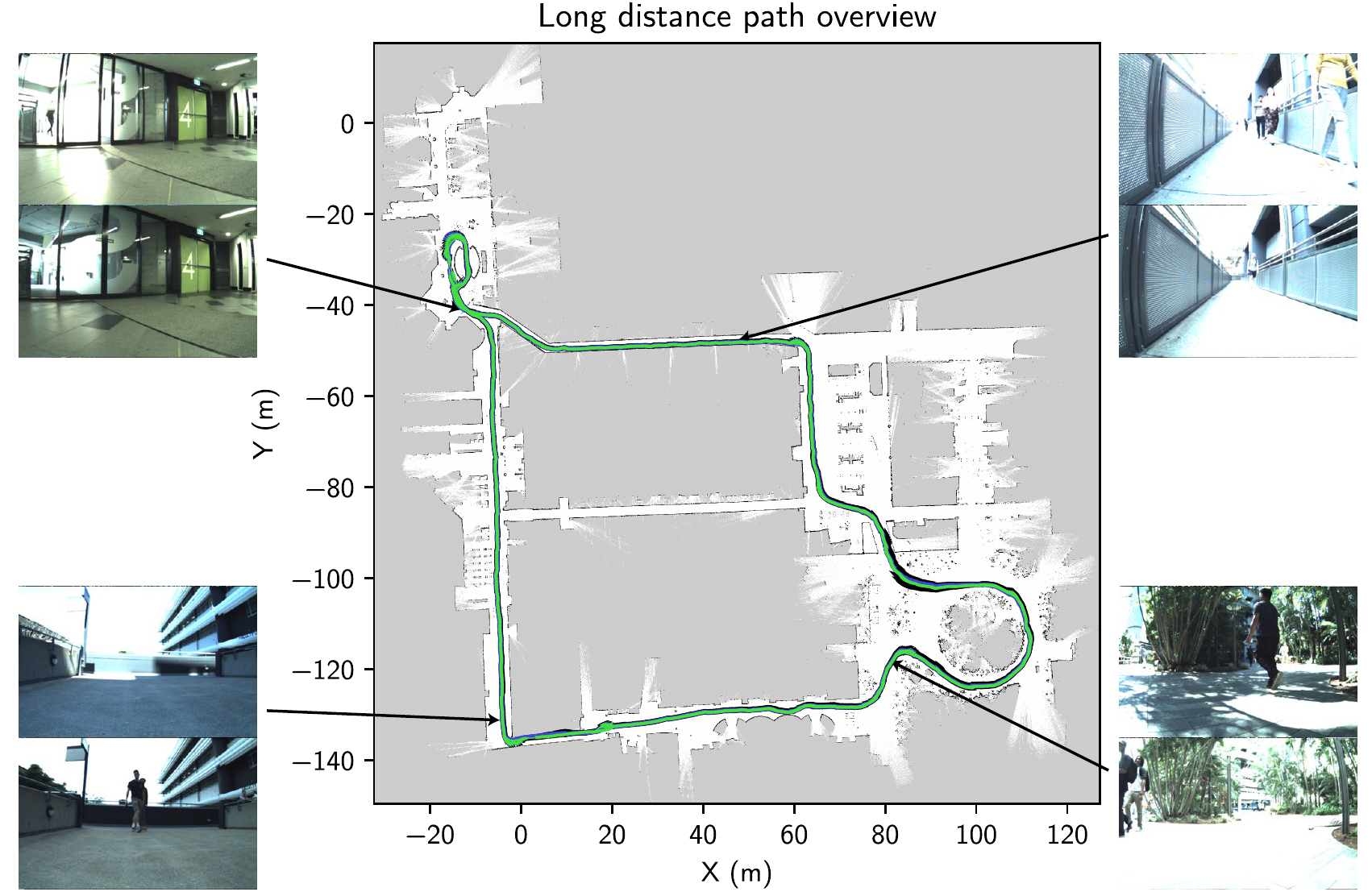}
	\vspace*{-0.1cm}
	\caption{Overview of the system's 550~m outdoor traverses. The teach run is shown in green, repeat runs in black, and the low resolution repeat run in blue. All repeat runs were successful, and closely align with the teach run. Example images are shown from the teach and repeat runs to illustrate challenging condition differences.}
	\vspace*{-0.25cm}
    \label{fig:outdoor-runs}
\end{figure}

A 550~m teach run was performed outdoors on the QUT university campus during bright sunny conditions. An overview of the run is shown in Fig.~\ref{fig:outdoor-runs}. Challenges included high contrast shadows in the environment and people walking past the robot. Successful repeat runs were performed with delays of one and four months after teaching, and with a reduced image resolution of $57\times22$.

IBVS~\cite{bistaAppearancebasedIndoorNavigation2016a} is specifically designed for indoor environments, and failed outdoors due to a lack of matching line features between the teach and repeat traverses, even when visual conditions were similar.
We also tested Bearnav~\cite{krajnikNavigationLocalisationReliable2018} on the outdoor route, which failed when lighting conditions differed between the teach and repeat runs.

\subsection{Extension Test 2: Transferring Teach Run Between Robots}
\label{subsec:transferteach}

For this test, we recorded a teach run on the Miro platform along the same indoor route as in Fig.~\ref{fig:indoor-runs}. As shown in Fig.~\ref{fig:robots}, Miro is much lower to the ground than the Jackal and has a wider field of view. Miro's odometry information is also much noisier than the Jackal's, with a noticeable leftward bias. The teach run was copied to the Jackal, with the only modification that the images were cropped to match the Jackal's field of view. Parameters were kept the same for the Jackal and Miro, as in Table~\ref{tab:params}; only the controller speed parameters were adjusted to account for each platform's dynamics.

Despite a different camera with perspective shift and different odometry between the two robots, the Jackal successfully repeated the run taught to Miro. While previous teach repeat approaches have operated on different robotic platforms, such as in\cite{bistaCombiningLineSegments2017,krajnikNavigationLocalisationReliable2018}, to the best of the authors' knowledge this is the first instance of a teach run being performed on one robotic platform and repeated on another. This demonstrates the high robustness of our approach to viewpoint change and odometry information, and provides promise that the approach could be easily deployed on other robots as further discussed in Section~\ref{sec:conclusion}.

\subsection{Computational Analysis}
Table~\ref{tab:computation} shows a computational comparison between methods for the default configurations. It is emphasised that our approach outperformed the state-of-the-art for the analysed configurations, while requiring less computation.

The main computational bottleneck of the approach is correlating the current image with each image in the search range, a process that could easily be parallelised. Real time (15 Hz) performance was achieved onboard the Miro, which has a Raspberry Pi 3B+ computer, by downscaling the images and reducing the along-path search range. 

\begin{table}[!t]
    \vspace*{0.2cm}
    \centering
    \footnotesize
    \caption{Computation comparison between approaches}
    \vspace*{-0.15cm}
    \label{tab:computation}
    \resizebox{0.99\linewidth}{!}{
    \begin{tabular}{c|c|c}
        Approach\Tstrut\Bstrut & Image processing & Full correction\\
        \hline
        Jackal ours: $115\times 44~(\alongpathsearchrange=3)$\Tstrut\Bstrut & 16~ms & 21~ms\\
        Miro ours: $29\times 11~(\alongpathsearchrange=1)$\Tstrut\Bstrut & 26~ms & 73~ms\\
        Jackal Bearnav~\cite{krajnikNavigationLocalisationReliable2018}: $752\times 480$\Tstrut\Bstrut & 25~ms & 28~ms\\
        Jackal IBVS~\cite{bistaAppearancebasedIndoorNavigation2016a}: $752\times 480$\Tstrut\Bstrut & 26~ms & 27~ms
    \end{tabular}
    }
    \vspace*{-0.5cm}
\end{table}

%% file: tex/5-conclusion.tex
\section{Conclusion}
\label{sec:conclusion}
Visual teach and repeat has been an active research topic over the past two decades, with applications ranging from a robot tour guide to interplanetary rovers. In this paper, we presented a novel teach and repeat approach that is particularly well suited for low-cost robots, as it can operate with a low-resolution monocular camera and noisy odometry information. This approach efficiently utilises odometry information for navigation, while employing a periodic vision-based correction signal to remain robust to odometry errors, leading to a very flexible system that can be easily deployed in practice. We have demonstrated that our approach works indoors and outdoors on different robotic platforms without parameter adjustment -- even when teach and repeat runs are executed on different robotic platforms. Detailed analysis has shown our approach outperforms the state-of-the-art in challenging situations including with noisy odometry or low resolution images, while requiring less computation.

Teach and repeat approaches are challenged in environments that have few nearby visual features, like large open fields, because the correction signal magnitude is reduced. Our approach's accuracy scales with the environment's visual density -- providing accuracy when needed. The system could be augmented to also focus on ground textures, allowing greater accuracy in a wider range of environments.
Additionally, our approach assumes that position or orientation errors cause predominantly horizontal image offsets, which could be violated in significantly non-flat environments or when using a distorted fish-eye camera. A small incline was successfully navigated in the outdoor route, and the distorted composite images from the Miro robot's two cameras did not impair navigation. But a natural extension would be to run a second stage of searching, such as vertical sweeping, when the horizontal match correlation is too low.

One further extension is to track the system's confidence in its path following and trigger a recovery manoeuvre if this confidence drops too low. There is also potential for offline preprocessing of the teach run images to improve matching of salient image features such as obstacles. Furthermore, we hope to use our system to cross-train navigation capabilities for different robots: for example, using reinforcement learning to train a navigation policy on the Jackal, and then repeating the learned policy on Miro. This is conceptually similar to the sim-to-real transfer problem, where it is well known that transferring policies is extremely challenging~\cite{ranamultiplicative,Nguyen2018}.